\documentclass[letterpaper, 10 pt, conference]{ieeeconf}  

\IEEEoverridecommandlockouts                              

\title{\LARGE \bf
Sample-Efficient Unsupervised Policy Cloning from Ensemble Self-Supervised Labeled Videos
}

\author{Xin Liu$^{1,2}$, Yaran Chen$^{3}$\textsuperscript{\Letter}, and Haoran Li$^{1,2}$\textsuperscript{\Letter} 
\thanks{This work was supported in part by the National Natural Science Foundation of China (NSFC) under Grants No.62136008 and No.62293545, and in part by the Excellent Youth Program of State Key Laboratory of Multimodal Artificial Intelligence Systems.}
\thanks{$^{1}$State Key Laboratory
of Multimodal Artificial Intelligence Systems, Institute of Automation, Chinese Academy of Sciences, Beijing 100190, China (emails: liuxin2021@ia.ac.cn, lihaoran2015@ia.ac.cn)}%
\thanks{$^{2}$School of
Artificial Intelligence, University of Chinese Academy of Sciences, Beijing
100049, China
        }%
\thanks{$^{3}$Department of Intelligent Science, Xi'an Jiaotong-Liverpool University, Suzhou 215123, China (email: yaran.chen@xjtlu.edu.cn)
        }%
\thanks{\textsuperscript{\Letter} Corresponding authors: Yaran Chen and Haoran Li
        }%
}

\usepackage{graphicx}
\usepackage{algorithmic}
\usepackage{algorithm}
\usepackage{amsmath,amsfonts}
\usepackage{booktabs}
\usepackage{multirow}
\usepackage{marvosym}
\usepackage{balance}

\makeatletter
\let\NAT@parse\undefined
\makeatother
\usepackage{hyperref}
\usepackage{cite}
\begin{document}

\maketitle
\thispagestyle{empty}
\pagestyle{empty}

\begin{abstract}

Current advanced policy learning methodologies have demonstrated the ability to develop expert-level strategies when provided enough information. However, their requirements, including task-specific rewards, action-labeled expert trajectories, and huge environmental interactions, can be expensive or even unavailable in many scenarios. In contrast, humans can efficiently acquire skills within a few trials and errors by imitating easily accessible internet videos, in the absence of any other supervision. In this paper, we try to let machines replicate this efficient watching-and-learning process through Unsupervised Policy from Ensemble Self-supervised labeled Videos (UPESV), a novel framework to efficiently learn policies from action-free videos without rewards and any other expert supervision. UPESV trains a video labeling model to infer the expert actions in expert videos through several organically combined self-supervised tasks. Each task performs its duties, and they together enable the model to make full use of both action-free videos and reward-free interactions for robust dynamics understanding and advanced action prediction. Simultaneously, UPESV clones a policy from the labeled expert videos, in turn collecting environmental interactions for self-supervised tasks. After a sample-efficient, unsupervised, and iterative training process, UPESV obtains an advanced policy based on a robust video labeling model. Extensive experiments in sixteen challenging procedurally generated environments demonstrate that the proposed UPESV achieves state-of-the-art interaction-limited policy learning performance (outperforming five current advanced baselines on 12/16 tasks) without exposure to any other supervision except for videos. 

\end{abstract}

\section{INTRODUCTION}

With the advancement of reinforcement learning (RL) \cite{atari-rl,continuous-rl,comsd,laskin2021urlb}, it's possible to train policies that reach expert levels for challenging tasks defined in complex environments \cite{icrassl,taco,liuguisong,haoran}. However, the demanding training conditions and low sample efficiency significantly constrain the applicability of RL. 
To this end, many advanced RL approaches, such as model-based RL \cite{simple,dreamer,protocad,dreamerv2} and self-supervised RL \cite{spr,earlysslrl,mind,lfs}, are proposed for higher sample efficiency. At the same time, some other researchers are no longer limited to online agent experience but try to seek help from offline supervisions. Given enough expert-labeled demonstrations, ideal policies can be obtained through offline RL \cite{offlinerl2,offline-rl1} or imitation learning \cite{il-survey,GAIL}, without the need for reward-specific environmental interactions. However, similar to expert rewards, these offline supervisions are also not widely available and always require time-consuming collection at a high cost. Different from them, the image-only videos are easily accessible currently thanks to the development of video websites and social media. 
However, Learning From Videos (LFV) poses a big challenge due to the lack of action labels. Although some attempts, such as video-based RL pre-training \cite{offlinerlpretrain, ICVF} and video-based intrinsic rewards \cite{diffusionreward}, have made great progress on RL data efficiency, they still require more or less additional supervisory information, e.g., reward-based finetuning and expert-annotated actions. Inverse RL methods \cite{GAIFO,LFV-IRL1,viper} provide effective solutions for completely unsupervised policy learning from videos by reward prediction. While they often fail to well balance policy performance and sample efficiency, requiring more interactions than RL with expert rewards \cite{ilpo-mp,il-survey}.

\begin{figure}[t]
    \centering
    \includegraphics[width=0.43\textwidth]{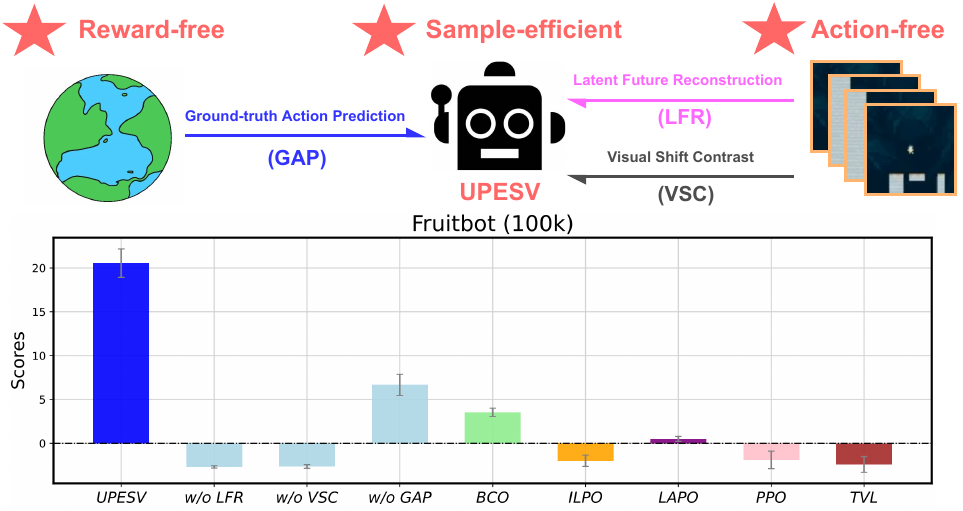}
    \vspace{-2mm}
    \caption{UPESV achieves sample-efficient policy learning with only action-free videos and reward-free interactions. This is achieved by three organically combined self-supervised tasks, where each performs its duties. For example, on the interaction-limited \textit{Fruitbot} task, UPESV obtains the only effective policy, where each self-supervised task is necessary. }
    \label{headimage}
    \vspace{-3mm}
\end{figure}

Can we derive a video-based policy with only a few interactions, without exposure to any other supervision? Previous advanced methods \cite{bco,ilpo} have proven it possible and achieved considerable results in state-based control (with the state-based demonstrations) and style-unchanged easy video games. However, their performance is unsatisfactory when facing environments with complex visual input and dynamics, which may be attributed to the following issues: They don't make full use of available information, i.e., both the action-free expert videos and reward-free interactions. For example, BCO \cite{bco} tries to label expert videos only based on non-expert environmental interactions, which means the highly qualified dynamics information contained in expert videos and the distribution difference are ignored. In contrast, ILPO \cite{ilpo} pays much more attention to the above dynamics contained in videos, but their world model is trained without considering the potential assistance provided by ground-truth interactions. In addition, their models are not provided with enough representation-specific signals. These signals are crucial for the models to well understand complex high-dimensional inputs \cite{yupeng,protorl,icrassl2}, thus extracting extensive action-related information (e.g., environmental dynamics) from both datasets.

In response to the above issues, we propose a novel, sample-efficient, and unsupervised video-based policy learning framework, named Unsupervised Policy from Ensemble Self-supervised labeled Videos (UPESV), as shown in Fig. \ref{headimage}. It organically employs several self-supervised tasks to well utilize both action-free expert videos and reward-free interactions, training a robust and dynamics-aware video labeling model to infer the expert actions for efficient policy cloning. 
For action-free videos, we employed a latent reconstruction task, where another world model is set to recover the next observation based on the current observation and predicted actions, in the latent space. This task forces the labeling model to understand the changes between neighboring observations and extract the high-quality dynamics information contained in expert videos, which is highly related to expert action prediction. For reward-free interactions, we let the labeling model predict the non-expert ground-truth actions. This task not only aligns the model to the real action space but also serves as a generalization task by changing both the data distribution and optimization target, thereby improving the model's robustness. In practice, we observe that naively gathering the above tasks into our framework can't yield effective policies in some environments, which can be mainly attributed to the unqualified representation learning. This motivates us to design a visual shift contrast task, where the representation module is required to associate two shifted images with the same origin. This task let the video labeling model pay more attention to action-related visual changes, enhancing its visual understanding ability in a target manner to well support the above two tasks. 
The organically employed three self-supervised tasks together enable advanced labeling performance, where each performs its duties. Along with the labeling model training, UPESV simultaneously clones a policy from the labeled expert videos, and collects environmental interactions to enrich self-supervised task data in turn. After a sample-efficient, unsupervised, and iterative training process, UPESV obtains an advanced policy based on a robust video labeling model. 

We conduct extensive experiments in sixteen challenging procedurally generated environments with complex visual inputs. Results demonstrate that the proposed UPESV achieves state-of-the-art policy learning under limited interactions, outperforming current advanced baselines on 12/16 tasks without exposure to any other supervision except for videos. 

We summarize the contributions as follows:
\begin{itemize}
    \item We propose a novel, sample-efficient, and reward-free framework to learn from action-free videos, named Unsupervised Policy from Ensemble Self-supervised labeled Videos (UPESV). Given a few environmental interactions, UPESV can derive an effective video-based policy without the need for any other supervision.
    \item UPESV trains a video labeling model through three organically combined self-supervised tasks while imitating a policy by cloning the labeled videos. Each task performs its duties, and they together provide the video labeling model with comprehensive dynamics understanding and advanced action prediction.
    \item Extensive experiments in sixteen challenging procedurally generated environments demonstrate that UPESV exhibits superior sample efficiency of policy learning. It outperforms five current advanced baselines on 12/16 tasks with limited interactions. 
\end{itemize}

\section{Related Works}
\subsection{Sample-efficient Policy Learning}

Interacting with the environment to collect data for policy learning is a time-consuming and expensive process. In some special scenarios, such as autonomous driving \cite{driving} and real robot learning \cite{realrobot}, the interactions can even lead to danger. To this end, many advanced methods are proposed to improve the RL sample efficiency. Model-based RL methods \cite{simple,dreamer,dreamerv3,dreamerpro} train extra world models that augment the RL experience, thereby improving sampling efficiency \cite{mbrl-survey}. Self-supervised RL, i.e., RL auxiliary tasks \cite{curl,mvp,vcr,crptpro} alleviate the lack of supervision signals for representation learning in DRL, accelerating policy learning by improving the upstream representation understanding capabilities. At the same time, many researchers try to completely avoid accessing environments, training policies through imitation learning \cite{il-survey,il-survey2} and offline RL \cite{offline-rl1,offlinerl2,offlinerlpretrain}. However, the required expert supervision signals, including both expert reward and offline expert dataset, are also expensive and unavailable in many scenarios. Different from these expert supervisions, videos are easy to obtain from the internet currently. Our work aims to achieve completely unsupervised policy learning with the easily accessible action-free videos, in the absence of rewards and any other expert supervision.

\begin{figure*}[t]
    \centering
    \includegraphics[width=0.83\textwidth]{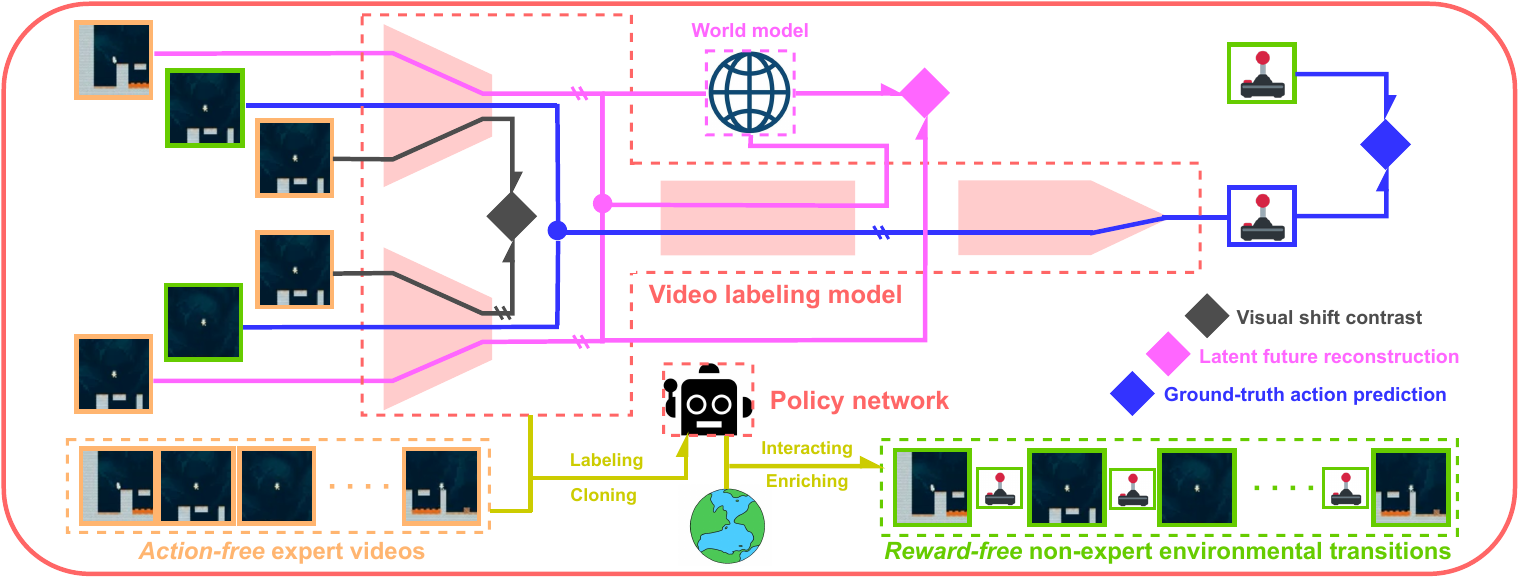}
    \vspace{-2mm}
    \caption{UPESV framework. UPESV learns a video labeling model and a policy network jointly through three organically combined self-supervised tasks, where each is necessary and performs its own duties. The motivation and details of these three tasks (visual shift contrast, latent future reconstruction, and ground-truth action prediction) are provided separately in Sections III.A, III.B, and III.C. Simultaneously, we imitate a policy $\pi(a|o)$ by behavior cloning the labeled expert videos. The policy interacts with the reward-free environment, in turn enriching the self-supervised training data for the video labeling model. The labeling model and policy are improved iteratively, which we detail in Section III.D.  }
    \label{diagram}
    \vspace{-3mm}
\end{figure*}

\subsection{Learning Policies From Videos}

Thanks to the improvement of the internet, it's effortless to acquire massive and different kinds of videos currently. Many videos contain expert demonstrations in different fields, providing extensive knowledge for both humans and machines. However, due to the lack of action information, how to well utilize the videos poses a big challenge. One of the most direct ways is to improve existing strategy learning methods with the help of videos, such as video-based pre-training for RL \cite{APV-LFV-reward1,VPT-LFV-ACTION1,mwm,mmwm} and video-based intrinsic reward for RL \cite{diffusionreward}. They indeed improve the sample efficiency of policy learning but still require expert supervision (i.e., RL rewards). Imitation Learning from only Observations (ILO) \cite{GAIFO,bco,ilpo} is proposed to completely decouple expert supervision from LFV. Most ILO methods try to extract expert rewards contained in the videos, employing RL to optimize the unsupervised reward expectation for policy imitation \cite{GAIFO,IRL-DMC1,IRL-DMC2}. These inverse RL methods achieve considerable results on imitation performance but can't take sample efficiency into account. At the same time, some researchers give up reward engineering for RL. By predicting expert actions from videos based on non-expert interactions \cite{bco,GSP} or digging out expert dynamics hidden in the expert videos \cite{ilpo,ilpo-mp}, they achieve more efficient unsupervised policy learning without expert actions. Different from them, UPESV benefits from both non-expert interactions and expert videos by organically employing multiple self-supervised tasks, achieving efficient unsupervised policy cloning for complex video games.

\section{Methodology}


UPESV trains a policy network and a video labeling model jointly. The policy determines the action $a$ based on the observation $o$, defined as $\pi(a|o)$. The video labeling model predicts the real expert action between two neighboring expert observations in videos, formulated as an inverse dynamics model $V(a^v_i | o^v_i , o^v_{i+1})$, where $a^v_i$ denotes the predicted action while $o^v_i$ and $o^v_{i+1}$ denote neighboring expert observations. $V$ consists of three networks: a feature encoder $f$, a latent predictor $g$, and an action projector $h$. $o^v_i$ and $o^v_{i+1}$ are separately encoded by $f$, concatenated, and processed by $g$ and $h$ to obtain the $a^v_i$, formulated as:
\begin{equation}
    a^v_i = h(g(f(o^v_i),f(o^v_{i+1}))).
\end{equation}

In practice, we also include a few available historical observations $o^v_{h}$ as additional input information. They are processed similarly to $o^v_i$ and we omit them in the remaining part for clarity. UPESV learns $V$ through three different but related self-supervised tasks over both expert videos $B^v$ and environmental interactions $B^e$, which we describe in the first three subsections. During the training of $V$, we simultaneously train the policy $\pi(a|o)$ by behavior cloning $V$-labeled expert videos, which we describe in Section III.D. The policy $\pi(a|o)$ collects self-supervised data for $V$, while the improved $V$ further enhances the policy $\pi(a|o)$. After a sample-efficient and iterative training process, we obtain the well-trained $V$ and $\pi$, as shown in Fig. \ref{diagram}.

\subsection{Self-supervised Task: Visual Shift Contrast}
\subsubsection{Motivation}
To infer actions from videos, the video labeling model $V(a^v_i | o^v_i , o^v_{i+1})$ should be able to capture the difference between neighboring images. In most visual control domains, the relative difference is more related to decision-making than the absolute difference in visual inputs. Imagine two particles moving upward at different positions on the plane. They have different absolute positions, but the same relative position changes that correspond to their same actions. To this end, we want $V$ to focus more on relative position difference, which is achieved by driving it to ignore the absolute position difference. Therefore, a shift-matching task is employed, forcing the video labeling model to align two randomly shifted images with the same origin in the semantic space. This task enhances the visual understanding ability of $V$ in a target manner, which is necessary for $V$ to understand the environmental dynamics through another two tasks (detailed in the next two subsections) simultaneously.

\subsubsection{Implementation}

The visual shift contrast task forces the labeling model $V$ to align two randomly shifted images with the same origin in the latent semantic space. Concretely, a batch of observations $\{o^v_i\}_{i=1}^N$ is sampled from the expert video dataset $B^v$. Each observation is randomly shifted (up to $s$ units in any direction) twice to obtain two shifted images, $\hat{o}^v_i$ and $\check{o}^v_i$, where the two different hats upon ${o}^v_i$ denote two independent random shifts. They are separately encoded by $f$ and the momentum encoder $f'$ to obtain two latent features, $f(\hat{o}^v_i)$ and $f'(\check{o}^v_i)$. $f'$ is not learned by gradient descent. It is updated through Exponential Moving Average (EMA) \cite{ema}, similar to \cite{curl,atc}. Then, we define ($f(\hat{o}^v_i)$,$f'(\check{o}^v_i)$) as positive pairs while ($f(\hat{o}^v_i)$,$f'(\check{o}^v_j)$) as negative pairs. Contrastive learning \cite{simclr} is conducted on the whole batch to bring the positive pair closer while pulling the negative pair further away from each other. A contrastive model that measures the distance is defined as follows:

\begin{equation}
    G_c(f(\hat{o}^v_i),f'(\check{o}^v_i)) = W\frac{u(f(\hat{o}^v_i))\cdot f'(\check{o}^v_i)}{||u(f(\hat{o}^v_i))||\cdot||f'(\check{o}^v_i)||},
\end{equation}

\noindent where $W$ is a trainable matrix. $u(\cdot)$ is a Multi-Layer Perceptron (MLP), which doesn't change the dimension of the latent embedding. It is set to introduce asymmetry, thereby avoiding collapse to trivial solutions. Then, we define an InfoNCE loss \cite{infornce} as the objective of visual shift contrast:

\begin{equation}
\begin{aligned}
    \mathcal{L}_{VSC} = &\log G_c(f(\hat{o}^v_i),f'(\check{o}^v_i)) \\
    &-\log \frac{1}{N} \sum_{j=1}^N G_c(f(\hat{o}^v_i),f'(\check{o}^v_j)).
\end{aligned}
\end{equation}

\subsection{Self-supervised Task: Latent Future Reconstruction}
\subsubsection{Motivation}

To predict expert actions accurately, the video labeling model $V(a^v_i | o^v_i , o^v_{i+1})$ should be able to capture and understand which relative differences (between neighboring images) are similar and close, which requires it to understand the expert-level environmental dynamics hidden in the videos. To this end, we introduce a latent future reconstruction task into model training, where another world model is employed. The world model takes the latent feature of the current observation and $V$-predicted latent action as inputs, trying to reconstruct the latent feature of the next observation. By supporting the world model to reconstruct different future observations in the videos, the labeling model is forced to understand the dynamics and refine more useful information into its small-dimension outputs.

\subsubsection{Implementation}
 The latent future reconstruction task updates the video labeling model $V$ and an extra world model $G_w$ jointly. First, a batch of observation pairs $\{(o^v_i,o^v_{i+1})\}_{i=1}^N$ is sampled from the expert video dataset $B^v$. Each current observation $o^v_i$ is encoded by the $f$ to obtain $f(o^v_i)$, while each next observation $o^v_{i+1}$ is encoded by the $f$ to obtain $f(o^v_{i+1})$. Then, they are concatenated and further processed by latent predictor $g$ to produce the predicted latent action $z^v_i$. $z^v_i$ is discretized by vector quantization \cite{vq} (the variable name is not changed for simplicity). Then, the $f(o^v_i)$ and $z^v_i$ are concatenated as the input to the world model $G_w(\cdot|z^v_i,f(o^v_i))$, and we can obtain a predicted observation $p^v_{i+1}$ in the latent space. Then we minimize the difference between $p^v_{i+1}$ and the latent embeddings of the ground-truth next observation $f(o^v_{i+1})$, jointly updating the parameters of our video labeling model $V$ and the world model $G_w$. The gradient is stopped before $f$ in backpropagation. The learning objective of this task is defined as the following:

 \begin{equation}
    \mathcal{L}_{LFR} = \frac{1}{N}\sum_{i=1}^N||f(o^v_{i+1}) - p^v_{i+1} ||^2.
\end{equation}

\noindent

\subsection{Self-supervised Task: Ground-truth Action Prediction}
\subsubsection{Motivation}

If the video labeling model $V(a^v_i | o^v_i , o^v_{i+1})$ can understand the expert video well, it should also generalize well to demonstrations sampled by other policies. To this end, we train our policy $\pi(a|o)$ to imitate the $V$-labeled videos through behavior cloning and utilize it to interact with environments for real transitions collection. Then $V$ is updated by achieving a real action prediction task on the collected real transitions (random collection in the first round).  With the $V$ improved by learning the real-action-labeled transitions, the derived policy $\pi$ is also improved and in turn collects more action-labeled data of higher quality for the ground-truth action prediction. This task not only provides the $V$ with the information of real action spaces, but also serves as a generalization task to improve both the robustness and performance of $V$ by different data distribution and optimization targets. Since the other two self-supervised tasks don't require real transitions, they can be carried out before this task and environmental interactions to further improve the sample efficiency.

\subsubsection{Implementation}

First, a batch of transitions $\{(o^e_i ,a^e_i, o^e_{i+1})\}^M_{i=1}$ is sampled from the environmental transition buffer $B^e$. We feed each observation pair into our video labeling model $V$ and obtain the $V$-predicted action $a^{e,V}_i$. The difference between the predicted action vector $a^{e,V}_i$ and the ground-truth action $a^e_i$ (one-hot) is minimized. The gradient is stopped before $g$ to reduce burden. The learning objective of this task is formulated as the following:

\begin{equation}
    \mathcal{L}_{GAP} = -\frac{1}{M}\sum_{i=1}^M \sum_{c=1}^C (a^{e,V}_{i,(c)} loga^e_{i,(c)}),
\end{equation}

\noindent where the subscript $(c)$ denotes the $c$-th vector dimension.

\begin{table*}[t]
\centering
\caption{The interaction-limited policy learning performance comparison between UPESV and all baseline methods. Each method is only allowed 100k steps on each task, which requires methods to have high sample efficiency to obtain an effective policy. The proposed UPESV exhibits state-of-the-art performance when compared with both advanced LFV methods and RL methods.}
\renewcommand\arraystretch{1.03}
\vspace{-1mm}
\setlength{\tabcolsep}{4.5mm}{
\begin{tabular}{ccccccc}
\toprule
Task               & UPESV (ours)      & BCO \cite{bco}             & ILPO  \cite{ilpo}   & LAPO \cite{lapo}             & PPO  \cite{ppo}    & TVL \cite{tvl}     \\ \midrule
\textit{Bigfish}   & \textbf{30.5 ± 1.6} & 3.6 ± 3.7          & 0.8 ± 0.1  & 20.6 ± 0.7          & 0.9 ± 0.1  & 1.0 ± 0.1  \\
\textit{Maze}      & \textbf{9.7 ± 0.2}  & 7.4 ± 2.4          & 4.2 ± 0.3  & 9.6 ± 0.1           & 5.0 ± 0.7  & 4.6 ± 0.7  \\
\textit{Heist}     & \textbf{9.4 ± 0.3}  & 7.6 ± 1.9          & 6.7 ± 0.5  & \textbf{9.4 ± 0.3}  & 3.7 ± 0.2  & 3.4 ± 0.9  \\
\textit{Coinrun}   & \textbf{7.4 ± 0.2}  & 6.7 ± 0.9          & 3.7 ± 1.3  & 6.2 ± 0.4           & 4.1 ± 0.5  & 3.6 ± 1.0  \\
\textit{Plunder}   & 3.5 ± 0.7           & 4.2 ± 0.3          & 2.2 ± 1.3  & \textbf{4.8 ± 0.1}  & 4.4 ± 0.4  & 4.5 ± 0.4  \\
\textit{Dodgeball} & \textbf{9.1 ± 0.8}  & 5.4 ± 1.1          & 0.6 ± 0.1  & 5.9 ± 1.1           & 1.1 ± 0.2  & 0.6 ± 0.2  \\
\textit{Jumper}    & 6.6 ± 0.2           & 6.4 ± 0.3          & 3.1 ± 0.6  & \textbf{7.3 ± 0.2}           & 3.5 ± 0.7  & 3.2 ± 0.5  \\
\textit{Climber}   & \textbf{6.8 ± 0.6}  & 3.3 ± 0.2          & 3.4 ± 0.5  & 4.7 ± 0.3           & 2.2 ± 0.2  & 2.4 ± 0.5  \\
\textit{Fruitbot}  & \textbf{20.6 ± 1.6} & 3.5 ± 0.5          & -2.0 ± 0.7 & 0.5 ± 0.3           & -1.9 ± 1.0 & -2.4 ± 0.9 \\
\textit{Starpilot} & 15.0 ± 0.8          & 12.8 ± 13.9        & 0.5 ± 0.7  & \textbf{20.3 ± 1.6} & 2.6 ± 0.9  & 1.8 ± 0.6  \\
\textit{Ninja}     & \textbf{6.3 ± 0.3}  & 4.2 ± 1.1          & 2.2 ± 1.1  & 5.2 ± 0.1           & 3.4 ± 0.3  & 3.0 ± 0.3  \\
\textit{Miner}     & \textbf{9.3 ± 1.2}  & 5.8 ± 1.3          & 1.2 ± 0.4  & 6.7 ± 0.6           & 1.2 ± 0.2  & 1.2 ± 0.2  \\
\textit{Caveflyer} & 3.5 ± 0.6           & 2.8 ± 1.1          & 3.2 ± 0.3  & \textbf{3.9 ± 0.1}  & 3.0 ± 0.4  & 3.2 ± 0.6  \\
\textit{Leaper}    & \textbf{2.9 ± 0.3}  & 2.5 ± 0.5          & 2.6 ± 0.2  & 2.7 ± 0.2           & 2.6 ± 0.3  & 2.5 ± 0.4  \\
\textit{Chaser}    & \textbf{0.8 ± 0.1}  & \textbf{0.8 ± 0.0} & 0.7 ± 0.0  & \textbf{0.8 ± 0.0}  & 0.4 ± 0.2  & 0.6 ± 0.2  \\
\textit{Bossfight} & \textbf{2.0 ± 0.4}  & 0.4 ± 0.3          & 0.1 ± 0.0  & 0.3 ± 0.3           & 0.1 ± 0.1  & 0.2 ± 0.2  \\ \midrule
Mean score         & \textbf{9.0}      & 4.8              & 2.1      & 6.8               & 2.3      & 2.1    \\ \bottomrule
\end{tabular}}
\vspace{-5mm}
\end{table*}

\subsection{Unsupervised Policy Cloning}

As mentioned in Section III.C, our policy $\pi(a|o)$ and video labeling model $V(a^v_i | o^v_i , o^v_{i+1})$ are improved jointly. $\pi(a|o)$ imitates the $V$-labeled expert videos through behavior cloning and interacts with environments for self-supervised data collection. With $V$ improved by the $\pi$-collected training data, $\pi(a|o)$ is also further enhanced and in turn provides more and better self-supervised training data. After a sample-efficient and iterative training process, an effective $\pi(a|o)$ and an advanced video labeling model $V$ are both obtained. 

$\pi$ consists of three networks: a feature encoder $f^{\pi}$, a policy predictor $g^{\pi}$, and an action projector $h^{\pi}$. To reduce the training burden, we share the visual encoder and action projector of the labeling model $V$ with those of $\pi$, i.e., $f^{\pi}=f$ and $h^{\pi}=h$, only training $g^{\pi}$ here. For higher sampling and training efficiency, $g^{\pi}$ is trained after learning $f^{\pi}$ (i.e., $f$) on videos and before learning $h^{\pi}$ (i.e., $h$) with interactions. For every observation pair $(o^v_i, o^v_{i+1})$ from the video dataset $B^v$, we use $V$ to obtain the predicted latent action $z^v_i$. Based on the current observation $o^v_i$ and the policy $\pi$, we obtain the policy-derived latent action $z^{v,\pi}_i = g^{\pi}(f^{\pi}(o^v_i))$. Then we minimize the difference between $z^v_i$ and $z^{v,\pi}_i$:

\begin{equation}
    \mathcal{L}_{UPC} = \frac{1}{N}\sum_{i=1}^N||z^{v,\pi}_i - z^v_i ||^2.
\end{equation}

The gradient is stopped before $f^{\pi}$. The policy-derived real action $a^{v,\pi}_i$ can be obtained by $a^{v,\pi}_i= h^{\pi}(z^{v,\pi}_i)$.

\section{Experiments \& Analysis}


\subsection{Experimental Settings}

We evaluated our method on all sixteen environments introduced by the Procgen benchmark \cite{procgen}. Procgen provides a diverse set of procedurally generated video game environments, each with unique challenges, multiple dynamics, and changing visual styles. For example, in \textit{Maze} task, the screen color, map size, and maze content are all changing drastically, which places high demands on both representation learning and dynamics understanding of learning methods.

We compare our UPESV with five current advanced online policy learning baselines, including both Learning From Video (LFV) methods and pure RL methods: BCO \cite{bco}, ILPO \cite{ilpo}, LAPO \cite{lapo}, PPO \cite{ppo}, and TVL \cite{tvl}. Among them, the first two methods are current state-of-the-art Imitation Learning from only Observation (ILO) methods, requiring no other supervision like UPESV. They either train a video labeling model for behavior cloning only on environmental transitions (BCO) or extract environmental dynamics from only expert videos (ILPO), while UPESV differs from these ILO baselines, employing several self-supervised tasks to utilize both datasets simultaneously for an advanced video labeling model. LAPO is a state-of-the-art video pre-training method. The pre-trained latent policy is finetuned by rewards and actions in environmental interactions. Inverse RL methods also provide solutions to learn video-based policy online without other supervision, but their goal is to restore the expert reward as much as possible, which makes them difficult to surpass RL methods exposed to expert rewards \cite{IRL-DMC1,IRL-DMC2}, in terms of sample efficiency \cite{ilpo-mp}. Considering that they are generally not evaluated on Procgen, we directly employ TVL and PPO, two state-of-the-art RL methods on Procgen provided with expert rewards in comparison. Note that our UPESV doesn't access these expert rewards and any extra supervision.

We choose \textit{full} distributions of levels and \textit{easy} mode in Procgen throughout this paper. For each method, 100k steps of environmental interactions are allowed. The expert videos required by the video-based methods contain 8M steps. They are generated by well-trained RL-based policies, provided by \cite{lapo}. For methods requiring rewards, we allow them to access the expert reward provided by Procgen environments. UPESV employs IMPALA-CNN \cite{impala} as encoders and predictors, MLP as projectors, and U-NET \cite{unet} as the world model. These follow previous works \cite{ppo,lapo,tvl} that are currently advanced on Procgen. The number of parallel environments is $64$, and the update frequency is $64$. The update times are 3k for ground-truth action prediction, 50k for unsupervised policy cloning, and 60k for the other objectives. Adam \cite{adam} is chosen as the optimizer, with the batch size set to $128$ for expert videos and $512$ for environmental transitions. The learning rate is set to $3e-5$ for the visual shift contrast, $3e-4$ for the latent future reconstruction, $1e-3$ for the ground-truth action prediction, and $2e-4$ for unsupervised policy cloning. The maximum shift distance $s$ in visual shift contrast is $1$, and the EMA momentum is set to $0.05$. The results of UPESV are obtained through $5$ different runs.

\begin{figure*}[!t]
    \centering
    \includegraphics[width=0.95\textwidth]{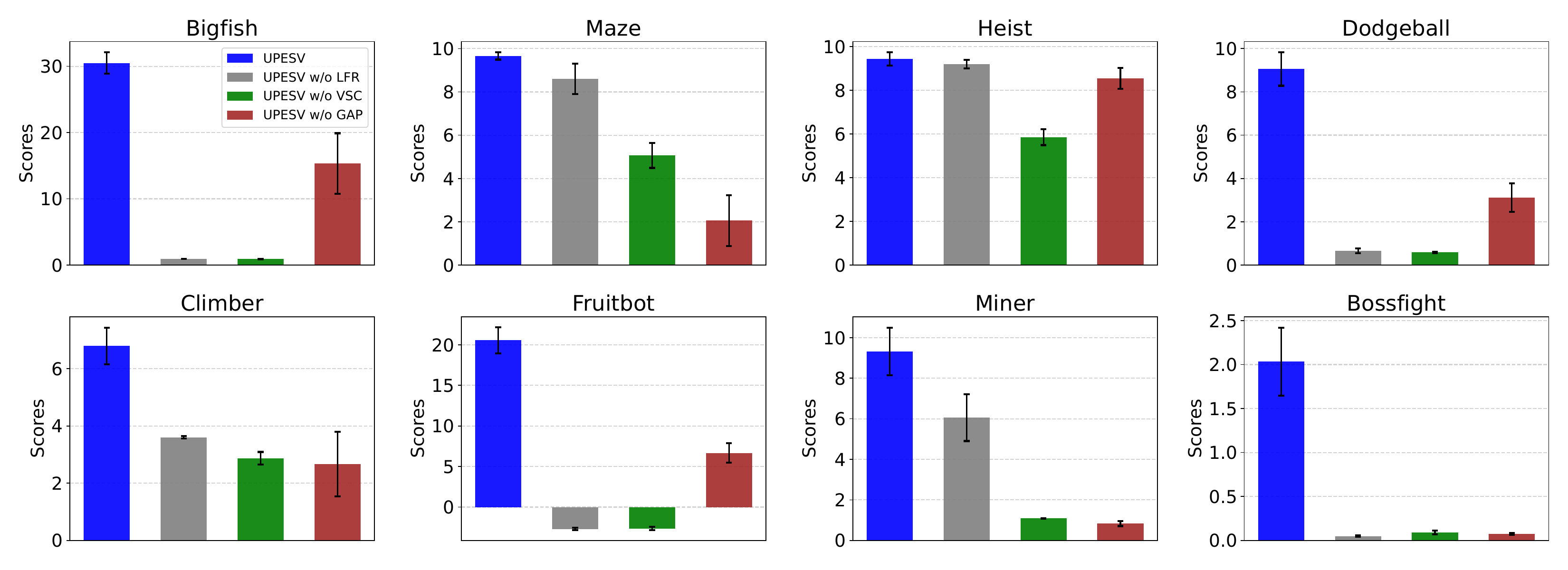}
    \vspace{-3mm}
    \caption{Ablation study of three self-supervised tasks on eight procgen environments. Each task is necessary in the proposed UPESV. }
    \vspace{-3mm}
    \label{ablation}
    
\end{figure*}

\begin{figure}[t]
    \centering
    \includegraphics[width=0.41\textwidth]{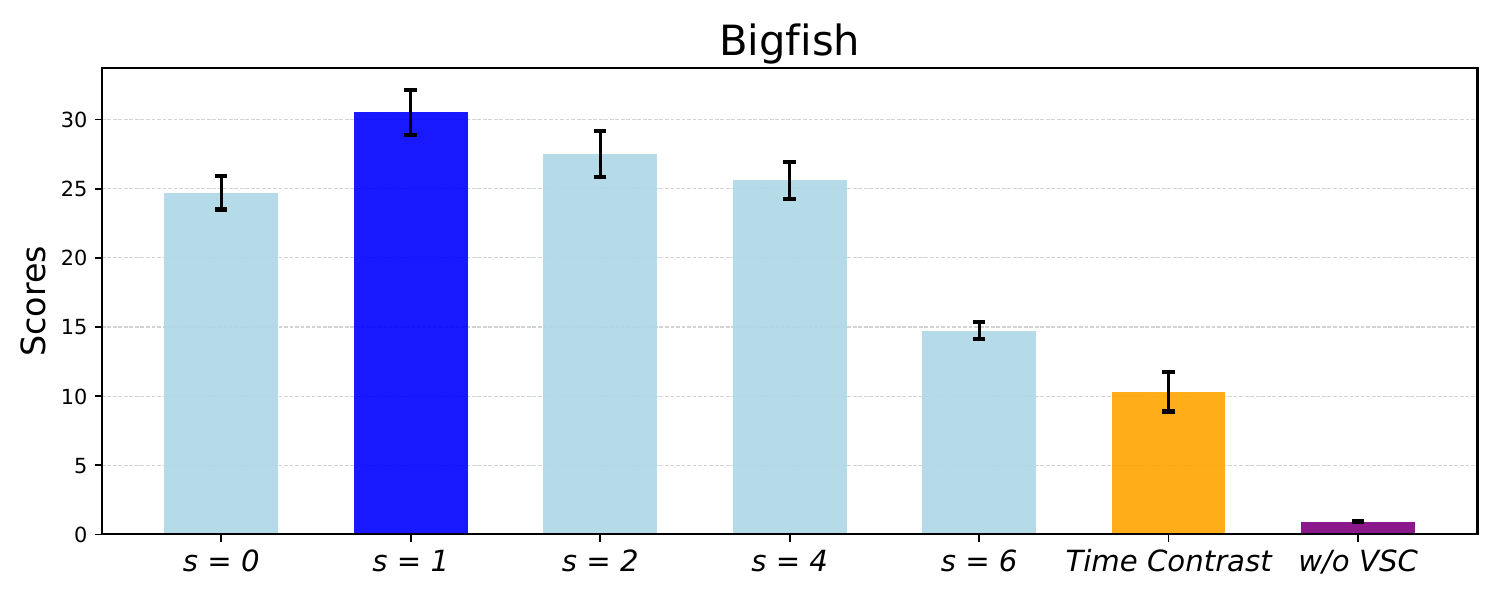}
    \vspace{-3mm}
    \caption{Hyper-parameter analysis of visual shift contrast. }
    \vspace{-3mm}
    \label{ssensity}
    
\end{figure}

\subsection{Sample Efficiency Comparison}

We test the policy learning abilities of all the methods when interaction is limited, which directly reflects their sample efficiency. Each method is allowed to interact with environments for only 100k steps. 100k steps are much less than the 25M employed in previous works \cite{procgen,tvl}, requiring extremely high sample efficiency. The results across all 16 tasks are shown in Table 1. Compared with three video-based baselines, UPESV performs better on 12/16 tasks and is also competitive on the others. These advantages demonstrate the UPESV video labeling model has a better understanding of both visual inputs and environmental dynamics, which can be attributed to the employed three self-supervised tasks. We observe trivial solutions that lead to performance drops in ILPO. This is due to its mode collapse in complex environments, which is also observed by \cite{ilpo-mp,lapo}. Compared with two advanced RL methods, UPESV's performance advantages are very obvious. In addition, video-based methods BCO and LAPO also overall perform better than pure RL methods. Note that both UPESV and BCO are reward-free. These phenomena demonstrate that the videos may serve as a more efficient teacher than expert rewards, especially when interactions are extremely limited. In summary, UPESV exhibits advanced sample efficiency without exposure to expert rewards or actions.

\subsection{Ablations: Necessity of Each Self-supervised Task}

In UPESV, we employ three different self-supervised tasks to jointly learn the video labeling model and policy. These three tasks are organically combined, where each one performs its own duties. To demonstrate that they are all indispensable in our framework, we sequentially ablate the latent future reconstruction task (UPESV w/o LFR), the ground-truth action prediction task (UPESV w/o GAP), and the visual shift contrast task (UPESV w/o VSC), observing the performance changes. In UPESV w/o GAP, we have to train the video labeling model and clone policies without accessing the true action space. To this end, we additionally train an action decoder for the cloned policy to remap the predicted latent actions to real actions based on environmental interactions and the world model, similar to \cite{ilpo}. The results are shown in Fig. \ref{ablation}, demonstrating that each employed self-supervised task is necessary for the proposed UPESV to achieve ideal policy cloning performance. 

\subsection{Hyper-parameter Sensitivity Analysis}

In this section, we test different shift distance $s$ values in the visual shift contrast (VSC) task, as shown in Fig. \ref{ssensity}. When $s=0$, VSC degenerated into an image discrimination task. It is also effective (much better than UPESV v/o VSC) but worse than adding a few shifts (worse than $s=1,2,4$. It means that forcing the model to ignore the global observation difference is helpful for understanding dynamics. In addition, we also try to add some time information (usually employed in RL), using two neighboring images instead of two shifted images (having the same origin) to define positive pairs for contrastive learning \cite{atc} while finding no improvements. It fits our intuition that the time contrast reduces the model's ability to perceive neighboring distinctions, which is critical for the understanding of complex environmental dynamics.

\begin{table}[t]
\centering
\caption{Prediction accuracy comparison on unseen expert dataset.}
\vspace{-2mm}
\renewcommand\arraystretch{1.1}
\setlength{\tabcolsep}{5mm}{
\begin{tabular}{ccc}
\toprule
Task      & UPESV (ours)    & BCO \cite{bco}  \\ \midrule
Starpilot & \textbf{50.4\%} & 32.6\% \\
Miner     & \textbf{33.1\%} & 10.4\% \\ \bottomrule
\end{tabular}}
\label{accuracy}
\vspace{-3.5mm}
\end{table}

\subsection{Robustness: Prediction Accuracy on Unseen Expert Data}

Considering that both BCO and our UPESV train video labeling models to infer real actions contained in expert videos, we compare the two models' prediction accuracy on unseen expert action-labeled datasets. The results in TABLE \ref{accuracy} show the superiority of UPESV. This is consistent with their performance on the policy learning tasks, demonstrating the influence of additional dynamics understanding and representation learning on models' robustness.

\section{Conclusion \& Limitation}

In this work, we propose UPESV, a novel, sample-efficient, and unsupervised framework to learn policies from action-free videos without rewards. With three organically selected and combined self-supervised tasks, UPESV enables state-of-the-art interaction-limited policy learning in multiple challenging procedurally generated video games. There are also some limitations. Although the design of ensemble self-supervised learning improves the sample efficiency, it inevitably results in a higher computational burden due to the additional objectives. In addition, balancing multiple self-supervised tasks with different learning rates also requires additional experimental trials. Finally, there is a huge gap between video games (discrete control) and continuous control or even real robots. How to bridge this gap is unsolved in this paper, and we leave it for future study.





\clearpage

\bibliographystyle{IEEEtran}
\balance
\bibliography{icra}

\end{document}